\documentclass[12pt]{iopart}
\usepackage{graphicx} 
\usepackage{iopams}
\usepackage[superscript]{cite}
\usepackage{setspace}
\usepackage{caption}

\begin{document}
\title[Uncertainty Quantification in Graph Neural Networks with Shallow Ensembles]{Uncertainty Quantification in Graph Neural Networks with Shallow Ensembles}

\author{Tirtha Vinchurkar$^1$, Kareem Abdelmaqsoud$^1$, John R. Kitchin$^1$\footnote{Corresponding author: \mailto{jkitchin@andrew.cmu.edu}}}

\address{$^1$  Department of Chemical Engineering, Carnegie Mellon University, 5000 Forbes Street, Pittsburgh, PA 15213, USA}

\ead{jkitchin@andrew.cmu.edu}

\begin{abstract}
Machine-learned potentials (MLPs) have revolutionized materials discovery by providing accurate and efficient predictions of molecular and material properties. Graph Neural Networks (GNNs) have emerged as a state-of-the-art approach due to their ability to capture complex atomic interactions. However, GNNs often produce unreliable predictions when encountering out-of-domain data and it is difficult to identify when that happens. To address this challenge, we explore Uncertainty Quantification (UQ) techniques, focusing on Direct Propagation of Shallow Ensembles (DPOSE) as a computationally efficient alternative to deep ensembles. By integrating DPOSE into the SchNet model, we assess its ability to provide reliable uncertainty estimates across diverse Density Functional Theory datasets, including QM9, OC20, and Gold Molecular Dynamics. Our findings often demonstrate that DPOSE successfully distinguishes between in-domain and out-of-domain samples, exhibiting higher uncertainty for unobserved molecule and material classes. This work highlights the potential of lightweight UQ methods in improving the robustness of GNN-based materials modeling and lays the foundation for future integration with active learning strategies.
\end{abstract}

\section{Introduction}

Machine-learned potentials (MLPs) have revolutionized materials discovery by providing accurate and efficient predictions of molecular and material properties \cite{fang_machine_2022, merchant_scaling_2023,xia_accelerating_2023}.
They are faster and more scalable than Density Functional Theory (DFT) calculations while maintaining comparable accuracy, making them ideal for large-scale materials simulations \cite{dragoni_achieving_2018,goeminne_dft-quality_2023,schleder_dft_2019}. Among these, Graph Neural Networks (GNNs) have emerged as state-of-the-art due to their ability to capture complex interactions within atomic structures. They outperform other machine learning models in predicting properties by learning graphical representations directly from molecular structures, rather than relying on predefined features \cite{fung_benchmarking_2021, louis_graph_2020, sunshine_chemical_2023}. However, like other MLPs, GNNs struggle in scenarios outside their training data. They often produce unreliable or incorrect predictions due to challenges in extrapolation, where the model's prediction uncertainty increases with distance from the training data points.

Such issues stem from sparse or biased training data, model limitations, and challenges in generalization. They could be attributed to sharp energy changes, new configurations, or localized phenomena, leading to errors during bond formation or structural changes \cite{varivoda_materials_2022}. This raises critical questions about the reliability of GNN-based predictions, especially in the field of materials discovery, where we aim to find new materials with exceptional properties, which usually correspond to the out-of-domain samples. To address this, Uncertainty Quantification (UQ) methods are essential for estimating confidence in predictions\cite{peterson_addressing_2017,singh_uncertainty_2021}. They help assess model robustness, identify out-of-domain applicability, and detect potential failure modes. Finally, the uncertainty estimation of these models can be integrated into active
learning algorithms for efficient exploration of the design space. \cite{xin_active-learning-based_2021}. 

Several methods have been proposed for Uncertainty Quantification (UQ) in machine learning models, each with unique advantages and challenges. Deep Ensembles \cite{egele_autodeuq_2022,rahaman_uncertainty_2021}, widely regarded as a benchmark for UQ, involve training multiple models independently with diverse initializations and aggregating their predictions to estimate uncertainty. These ensembles provide uncertainty estimates but they are computationally expensive due to the need for training multiple models. The Latent Distance method \cite{musielewicz_improved_2024}, on the other hand, leverages the distance in a model’s latent space to estimate uncertainty, effectively capturing data distribution shifts and out-of-domain inputs. However, its success depends heavily on the quality of the model’s latent space representation.

Bayesian Neural Networks (BNNs)\cite{olivier_bayesian_2021} offer a theoretically grounded approach by treating model weights as distributions rather than fixed parameters, thus incorporating uncertainty directly into the model. Despite their strong theoretical foundation, BNNs face scalability issues due to high computational costs and the complexity of approximating posterior distributions. Monte Carlo Dropout \cite{gal_dropout_2016} presents a more practical alternative by applying dropout during inference and sampling multiple predictions to estimate uncertainty. While it is computationally less demanding than BNNs, its uncertainty estimates can lack robustness in high-dimensional data or when dropout rates are not well-tuned. Ultimately, the choice of UQ method depends on the application’s computational constraints and the desired balance between accuracy and efficiency in uncertainty estimation.

Direct Propagation of Shallow Ensembles (DPOSE) \cite{kellner_uncertainty_2024} represents a promising alternative to those methods by addressing key limitations such as computational overhead and complexity. By using shallow ensembles with weight sharing and Negative Log-Likelihood (NLL) loss, DPOSE achieves reliable uncertainty estimates efficiently. Its lightweight, plug-and-play nature aligns with this study's objectives, making it an ideal choice for scalable and transparent uncertainty propagation across diverse models and datasets.

In this work, we utilize DPOSE to quantify the uncertainty of the SchNet model\cite{schutt_schnet_2018} on three diverse datasets: OC20\cite{chanussot_open_2021}, QM9\cite{ramakrishnan_quantum_2014,ruddigkeit_enumeration_2012}, and a Gold dataset\cite{boes_neural_2016} comprising various structural morphologies. Our results show that DPOSE tends to capture meaningful uncertainty trends across these datasets, providing valuable insights into both in-domain and out-of-domain samples. While the complexity of certain datasets presents challenges in clearly differentiating data domains, DPOSE demonstrates its potential as a scalable and efficient method for uncertainty estimation in large-scale applications.

\section{Methods}

An ideal uncertainty propagation method should be computationally efficient, easy to implement, transparent in its methodology, aligned with model errors, and scalable across datasets and models. Direct Propagation of Shallow Ensembles (DPOSE) \cite{kellner_uncertainty_2024} meets these criteria by offering a streamlined alternative to traditional deep ensembles while maintaining reliable uncertainty estimates.

DPOSE strikes an effective balance between computational efficiency and reliable uncertainty estimation. While direct mean-variance prediction methods are straightforward and computationally cheap, they often suffer from poor calibration, training instability, and limited generalization, particularly when dealing with out-of-domain data. In contrast, DPOSE leverages shallow ensembles with weight sharing to capture both aleatoric and epistemic uncertainties more robustly, providing better-calibrated and more stable uncertainty estimates. By preserving ensemble diversity without the high computational cost of deep ensembles, DPOSE mitigates issues such as model bias and overfitting inherent in single-model approaches. Its lightweight design, transparent methodology, and scalability across models and datasets make it particularly well-suited for large-scale materials modeling applications, where both accuracy and efficiency in uncertainty quantification are essential.

\begin{figure}
    \centering
    \includegraphics[width=\textwidth]{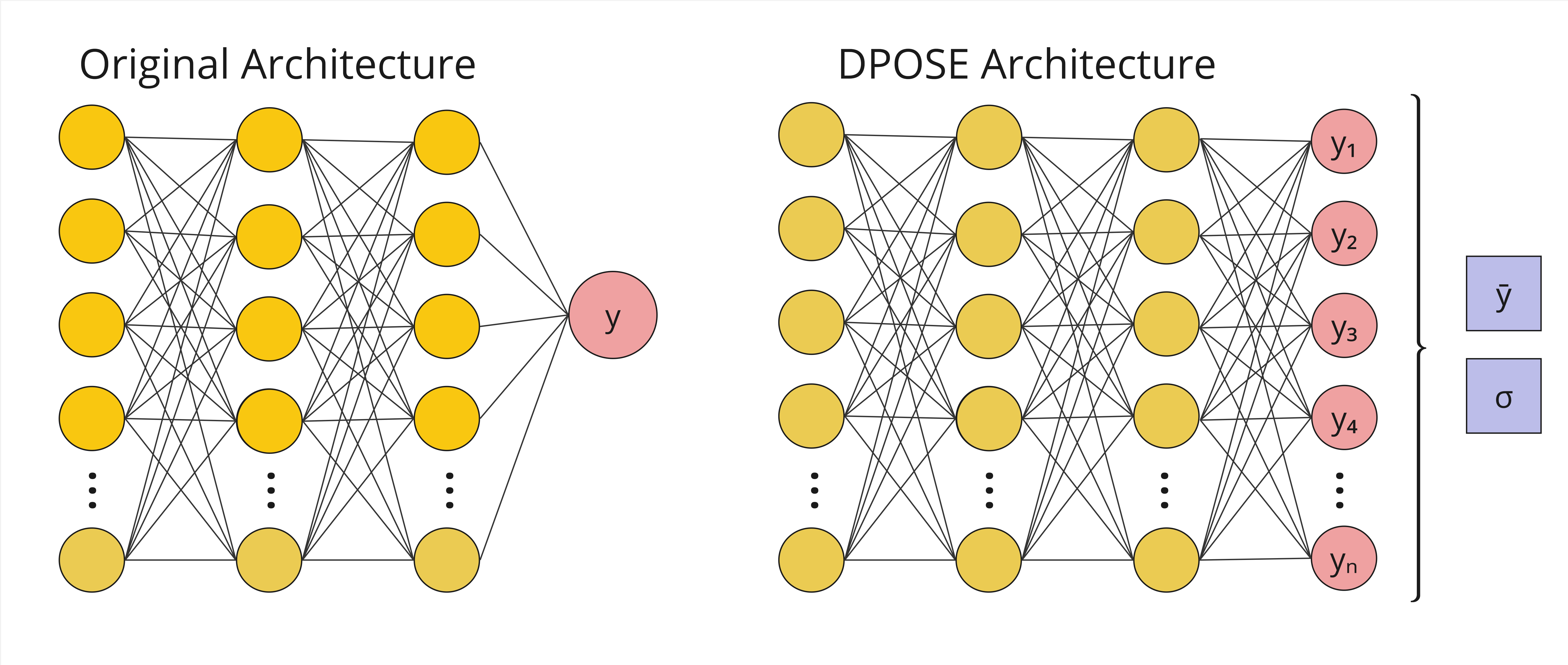}
    \caption{Comparison of the Original Architecture and the DPOSE Architecture. The Original Architecture produces a single output (\(y\)) using a deep neural network. The DPOSE Architecture consists of multiple shallow ensembles in the last layer predict outputs (\(y_1, y_2, \dots, y_n\)), with the final output (\(\bar{y}\)) computed as their average and the uncertainty (\(\sigma\)) estimated as their variance.}

    \label{fig:dpose_architecture}
\end{figure}

DPOSE modifies the final layer of a model (see Figure \ref{fig:dpose_architecture}) and employs Negative Log-Likelihood (NLL) loss (as shown in Equation \ref{eq:nll}), balancing both the Mean Square Error (MSE) Loss and variance terms. It uses shallow ensembles to approximate posterior distributions efficiently, enabling transparent and scalable uncertainty propagation. Key features include an ensemble architecture for predicting both mean and variance, weight sharing up to the last layer to reduce computational costs, and integrated error estimation for derived quantities. 

\begin{eqnarray}
  \mathrm{NLL}(\Delta y,\sigma)
    &=& \frac{1}{2}\Bigl[\frac{\Delta y^2}{\sigma^2}
         + \ln\bigl(2\pi\,\sigma^2\bigr)\Bigr]
  \label{eq:nll}
\end{eqnarray}

These characteristics make DPOSE compatible with most deep learning architectures, requiring only minimal changes to the final layer and loss function. Its computational cost is comparable to adding an extra hidden layer, making it a practical and effective solution for high-dimensional molecular systems.

\subsection{Model Selection and Datasets}
This study explores Uncertainty Quantification (UQ) for the SchNet model \cite{schutt_schnet_2018}, a widely used neural network for atomistic simulations. While newer models like DimeNet++\cite{gasteiger_fast_2022,gasteiger_directional_2022} or Equiformer \cite{liao_equiformer_2023,liao_equiformerv2_2024} may offer higher accuracy, SchNet strikes a good balance between computational efficiency and performance. By integrating DPOSE into SchNET, we aim to achieve a balance between model efficiency and reliable uncertainty estimation. 

We applied this approach to three datasets: QM9\cite{ramakrishnan_quantum_2014,ruddigkeit_enumeration_2012}, Open Catalyst 2020 (OC20)\cite{chanussot_open_2021}, and the Gold dataset \cite{boes_neural_2016}, chosen to test the model across diverse systems.  QM9, a benchmark dataset of small organic molecules, was included to evaluate how the model uncertainty varies for molecular systems. OC20, which focuses on heterogeneous catalytic surfaces with molecular adsorbates, was selected to test the model’s ability to quantify uncertainty in intermetallic and nonmetal slab systems where surface interactions dominate. Unlike QM9, this dataset incorporates periodic boundary conditions. The Gold dataset extends the evaluation to Molecular Dynamics and Simulation frames of Gold systems at both equilibrium and non-equilibrium states in different form such as bulk, clusters, and surfaces. This dataset allows us to examine how model variance adapts to both ordered atomic environments (e.g., bulk gold) and disordered environments (e.g., amorphous, clusters).

The objective of this study is to determine if the model can reliably differentiate between in-domain and out-of-domain data. The model is trained/fine-tuned on in-domain data and evaluated on both in-domain and out-of-domain datasets, with the goal of producing lower uncertainty estimates for in-domain samples and higher uncertainty for out-of-domain samples.

In this context, we consider in-domain samples to be those included in the fine-tuning dataset, enabling the model to learn their specific characteristics and produce confident predictions with low uncertainty. In contrast, out-of-domain samples are not part of the fine-tuning dataset. These samples may have been seen during the initial pretraining phase or be entirely novel but differ from the fine-tuning data in structure, composition, or other relevant features. As a result, the model is expected to assign higher uncertainty to out-of-domain samples, reflecting reduced confidence in unfamiliar regions.

\subsection{Experimental Setup}

The SchNet model was modified by splitting its final layer into 64 output heads. This enabled 64 independent predictions of the total energy, forming the SchNet Ensemble (SE) model. The mean and variance of these predictions were calculated and integrated into the Negative Log-Likelihood (NLL) loss function for training.
\subsubsection{QM9 dataset}
The QM9-SE model was trained on the entire QM9 dataset. The dataset includes 130,831 small organic molecules ($C_{1}$ to $C_{9}$) in their equilibrium states, containing elements C, H, O, N, and F. Training was conducted for 1000 epochs using an Adam optimizer, with a learning rate of $5 \times 10^{-5}$, a batch size of 100, and an 80:20 train-test split.

The evaluation was conducted across three scenarios to assess the model’s ability to quantify uncertainty and generalize to out-of-domain data:
\begin{enumerate}
    \item \textit{Equilibrium vs. Non-equilibrium States:} Molecules in equilibrium states (present in the dataset) were compared against non-equilibrium states (absent) to evaluate the model's ability to detect and handle configurations not seen during training. The non-equilibrium states were generated by systematically altering the bond length between two carbon atoms in the molecule, simulating stretched or compressed configurations by increasing the distance from the equilibrium state.
    \item \textit{Known vs. Unknown Elements:} Molecules containing only the dataset elements (C, H, O, N, F) were contrasted with those containing unseen inorganic elements (Si, S, P, Cl) to test the model’s response to out-of-domain compositions.
    \item \textit{Small vs. Larger Molecules:} The model’s predictions for small molecules ($C_1$ to $C_9$, present in the dataset) were compared with predictions for larger molecules ($C_{10}$ to $C_{18}$, absent from the dataset) to observe how uncertainty estimates scale with molecular size.
\end{enumerate}

These experiments were designed to systematically evaluate the SchNet model’s uncertainty quantification capabilities, particularly its ability to distinguish in-domain and out-of-domain data reliably.

\subsubsection{OC20 dataset}
This dataset contains over 1.2 million DFT relaxations spanning diverse materials, surfaces, and adsorbates. These include 82 molecular adsorbates and approximately 11,451 surfaces \cite{chanussot_open_2021}. To simplify the problem, intermetallic and nonmetal slabs were selected as the focal subset.  The OC20-SE model was developed by fine-tuning a pretrained SchNet model on the intermetallic slab subset to improve its specialization. This approach enables evaluation of the DPOSE method specifically on crystalline materials. The pretrained SchNet checkpoint was originally trained on the entire OC20 dataset. During fine-tuning, all weights were frozen except for the modified final layer.

The evaluation was conducted in two ways to assess the model’s uncertainty quantification and performance:
\begin{enumerate}
    \item \textit{Intermetallic vs. Non-metal Slabs:} The OC20-SE model's performance and variance for the in-domain dataset (metal slabs) was compared against the out-of-domain dataset (non-metal slabs). Low variance is expected for the in-domain dataset due to the model’s higher accuracy and confidence resulting from fine-tuning on intermetallic slabs. High variance is expected for the out-of-domain dataset, where the model encounters systems outside its training distribution, resulting in lower accuracy.
    \item \textit{Volume per atom Compression and Expansion:} For both in-domain and out-of-domain datasets, the model’s behavior was tested by compressing and expanding the volume per atom, which inherently alters the bond lengths between atoms. This approach provides a convenient way to generate non-equilibrium states. The potential energy was plotted against the corresponding variance. Since the OC20 dataset contains systems that are geometrically optimized and in equilibrium states, the expected behavior is low variance at the equilibrium state (minimum potential energy) and increasing variance as the system moves away from equilibrium.
\end{enumerate}

\subsubsection{Gold Dataset}  
The dataset includes gold systems in bulk, cluster, and surface configurations. Bulk structures are categorized into FCC, BCC, SC, HCP, and diamond cubic lattices, while amorphous systems belong to the cluster group. These frames span equilibrium and non-equilibrium states. For this study, we focused on comparing bulk systems (highly ordered lattices) with amorphous systems (disordered clusters), comprising 797 and 6,278 systems, respectively.  

The AuMD-SE model was developed by fine-tuning a SchNet model checkpoint pre-trained on the OC20 dataset. Fine-tuning was performed on bulk systems using the Negative Log-Likelihood (NLL) loss function. The training setup included a batch size of 20, an evaluation batch size of 8, 200 epochs, and a learning rate of \(1 \times 10^{-4}\).  

The evaluation was conducted to analyze model performance and uncertainty trends:  
\begin{enumerate}  
    \item \textit{Bulk vs. Amorphous Systems:} The AuMD-SE model's performance was compared between bulk (ordered) and amorphous (disordered) systems to assess uncertainty trends across different structural morphologies. The model was fine-tuned on bulk systems, which are highly ordered, and is therefore expected to produce low uncertainty for these configurations. In contrast, amorphous systems, characterized by their disordered structures and absent from the fine-tuning dataset, are expected to exhibit higher uncertainty.  
    \item \textit{Energy-Variance-Error analysis:}  The relationship between energy, variance, and error was examined by sorting the energy values and analyzing their corresponding variance and error. Since the dataset inherently contains both equilibrium and non-equilibrium states, perturbing the volume per atom (as done in prior tests) to compare equilibrium versus non-equilibrium configurations no longer provided meaningful distinctions. Instead, this direct, one-to-one comparison across energy ranges offered targeted insights into how the model’s predictions evolve with varying energy levels and revealed correlations between uncertainty (variance) and prediction accuracy (error). 
\end{enumerate}

\section{Results and Discussion}

\subsection{QM9 Dataset}

\subsubsection{Equilibrium and Non-equilibrium States of Molecules}
\begin{figure}
    \centering
    \includegraphics[width=\textwidth]{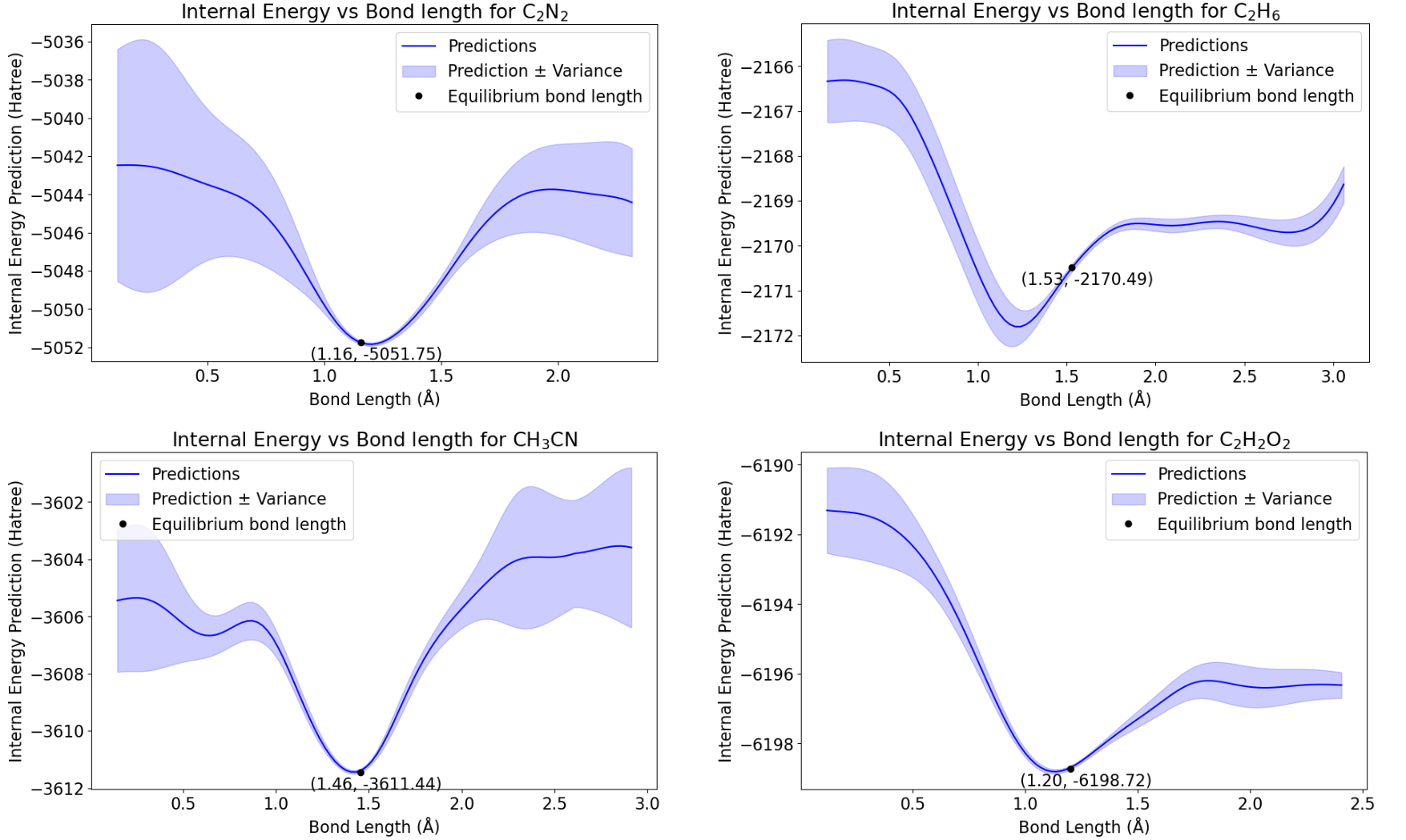}
    \caption{Internal energy predictions and uncertainty estimates for C$_2$N$_2$, C$_2$H$_6$, CH$_3$CN, and C$_2$H$_2$O$_2$ as a function of bond length between two carbon atoms. Solid lines represent predicted internal energies, and shaded regions indicate prediction variance. Minimum variance is observed at equilibrium bond lengths from the training dataset.}

    \label{fig:uncertainty_behavior}
\end{figure}

To assess the QM9-SE model’s behavior on equilibrium and non-equilibrium configurations, we selected four test cases: ethane (C$_2$H$_6$), cyanogen (C$_2$N$_2$), acetonitrile (CH$_3$CN), and glyoxal (C$_2$H$_2$O$_2$). The model was trained on equilibrium geometries, and for each molecule, the bond length between the two carbon atoms was systematically expanded and compressed. As the bond length deviated from equilibrium, the model's uncertainty (measured as variance) increased, while the lowest uncertainty was observed near the equilibrium point, as shown in Figure \ref{fig:uncertainty_behavior}. This behavior is consistent with the training data, which predominantly consists of equilibrium geometries, reflecting the model's limited confidence in extrapolated configurations.

\subsubsection{Inorganic Elements Absent in the QM9 Dataset} 
To evaluate the model's performance on molecules containing elements absent in the QM9 dataset, we analyzed a series of test cases, including both in-domain and out-of-domain molecules. The in-domain molecules, such as CF$_4$, CH$_4$, and HF, showed consistently low uncertainty due to their representation in the training data. In contrast, out-of-domain molecules with absent elements, such as Si and Cl, exhibited significantly higher uncertainties. For instance, replacing fluorine with chlorine in CF$_4$ (forming CCl$_4$) caused the variance to increase from 0.098 to 62. Similarly, substituting carbon with silicon in CH$_4$ (forming SiH$_4$) increased the variance from 0.093 to 2.7. Molecules entirely absent from the training data, such as SiCl$_4$, exhibited high uncertainty (56), further validating the model's ability to flag out-of-domain chemistry. These results, summarized in Table \ref{tab:inorganic}, demonstrate the framework's robustness in identifying molecular systems outside its training domain.

\begin{table}
\caption{\label{tab:inorganic}In-Domain and Out-of-Domain variances for QM9 molecules. The variance in the in-domain systems is lower than the out-of-domain systems as expected.}
\begin{indented}
\item[]\begin{tabular}{@{}llll}
\br
In-Domain & Variance & Out-Of-Domain & Variance \\
\mr
CF$_4$ & 0.09 & CCl$_4$ & 62 \\
CH$_4$ & 0.09 & SiH$_4$ & 2.7 \\
HF & 0.21 & HCl & 3.2 \\
& & SiF$_4$ & 1.7 \\
& & SiCl$_4$ & 56 \\
\br
\end{tabular}
\end{indented}
\end{table}

\subsubsection{Larger Molecules}
To analyze the model's uncertainty for larger molecular systems, we studied alkanes (C$_1$–C$_{18}$) and alcohols (C$_1$–C$_{18}$), measuring uncertainties across increasing chain lengths. In both chemical families, the uncertainty gradually increased as molecular size grew, reflecting the model's limited exposure to larger systems during training. For alkanes, uncertainties ranged from 0.092 for methane (C$_1$) to 0.66 for octadecane (C$_{18}$). A similar trend was observed for alcohols, with uncertainties increasing from 0.082 for methanol (C$_1$) to 0.70 for octadecanol (C$_{18}$). These findings suggest that the DPOSE framework captures the growing complexity and potential extrapolation error associated with larger molecules. Detailed results for alkanes and alcohols are provided in Table \ref{tab:alkanes}.

\begin{table}
\caption{\label{tab:alkanes}Alkane and Alcohol Uncertainty Comparison. The uncertainty gradually increases as molecular size grows, reflecting the model’s limited exposure to larger systems during training.}
\begin{indented}
\item[]\begin{tabular}{@{}llll}
\br
Alkanes & Variance & Alcohols & Variance \\
\mr
\textbf{Methane} & \textbf{0.09} & \textbf{Methanol} & \textbf{0.08} \\
Ethane & 0.09 & Ethanol & 0.07 \\
Propane & 0.07 & Propanol & 0.06 \\
Butane & 0.08 & Butanol & 0.06 \\
Pentane & 0.08 & Pentanol & 0.06 \\
Hexane & 0.08 & Hexanol & 0.07 \\
Heptane & 0.07 & Heptanol & 0.07 \\
Octane & 0.08 & Octanol & 0.09 \\
Nonane & 0.11 & Nonanol & 0.12 \\
Decane & 0.16 & Decanol & 0.18 \\
Undecane & 0.22 & Undecanol & 0.25 \\
Dodecane & 0.62 & Dodecanol & 0.38 \\
Tridecane & 0.82 & Tridecanol & 0.35 \\
Tetradecane & 0.61 & Tetradecanol & 0.44 \\
Pentadecane & 0.45 & Pentadecanol & 0.43 \\
Hexadecane & 0.68 & Hexadecanol & 0.83 \\
Heptadecane & 0.89 & Heptadecanol & 0.87 \\
\textbf{Octadecane} & \textbf{0.66} & \textbf{Octadecanol} & \textbf{0.70} \\
\br
\end{tabular}
\end{indented}
\end{table}

\subsection{OC20 Dataset}

\subsubsection{Intermetallic and Non-metal Slabs}
\begin{figure}
    \centering
    \includegraphics[width=0.9\textwidth]{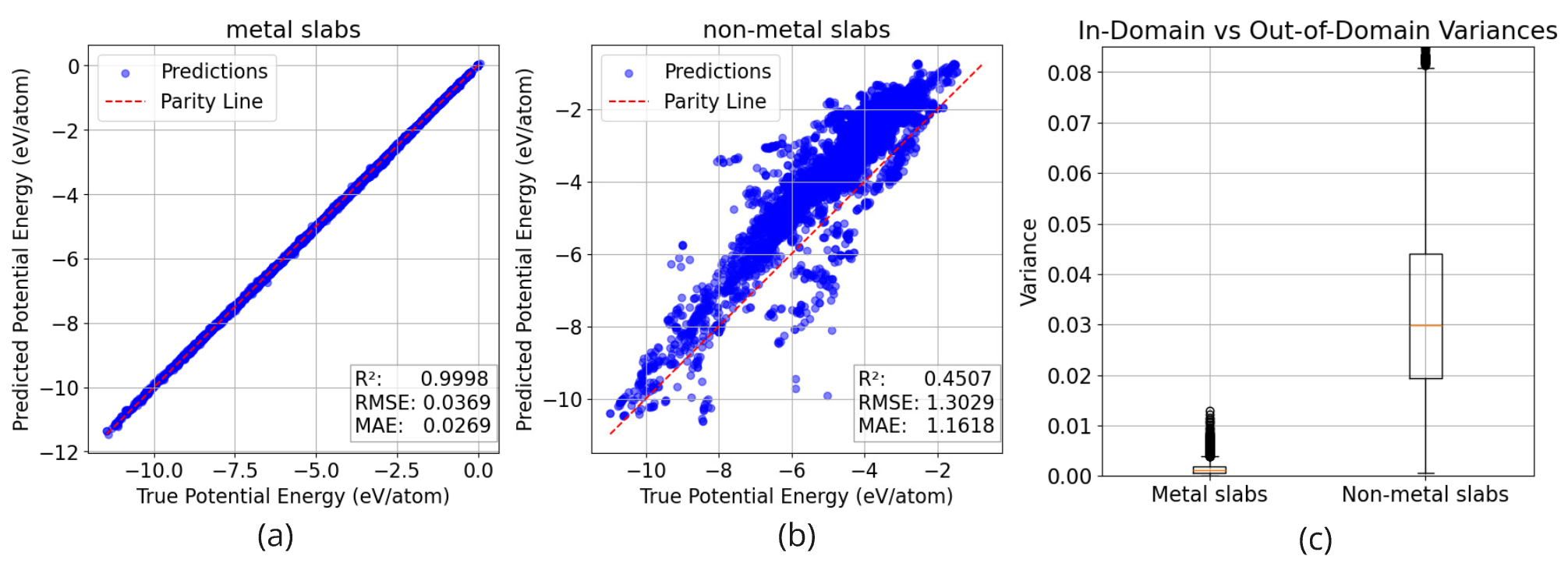}
    \caption{(a) Parity plot for inter-metallic slabs showing high accuracy.
    (b) Parity plot for non-metal slabs showing lower accuracy.
    (c) Variance estimates show higher uncertainty for non-metal slabs than inter-metallic slabs.}
    \label{fig:oc20_model_performance}
\end{figure}

The results indicate that the OC20-SE model achieves higher predictive accuracy for intermetallic slabs, with $R^2$ scores nearing 1 and lower Root Mean Square Error (RMSE) and Mean Absolute Error (MAE) values, as illustrated in Figure \ref{fig:oc20_model_performance}(a). In contrast, predictions for non-metal slabs exhibit larger deviations from the ground truth potential energy values, as shown in Figure \ref{fig:oc20_model_performance}(b). Additionally, variance estimates reveal greater uncertainty for non-metal slabs. The median variance for non-metal slabs is approximately 0.03, whereas the upper whisker of the variance for inter-metallic slabs remains below 0.01, as depicted in Figure \ref{fig:oc20_model_performance}(c). These findings underscore the model's accuracy toward the inter-metallic slab systems due to the fine-tuning process. 

\subsubsection{Change in Volume Per Atom}
\begin{figure}
    \centering
    \includegraphics[width=\textwidth]{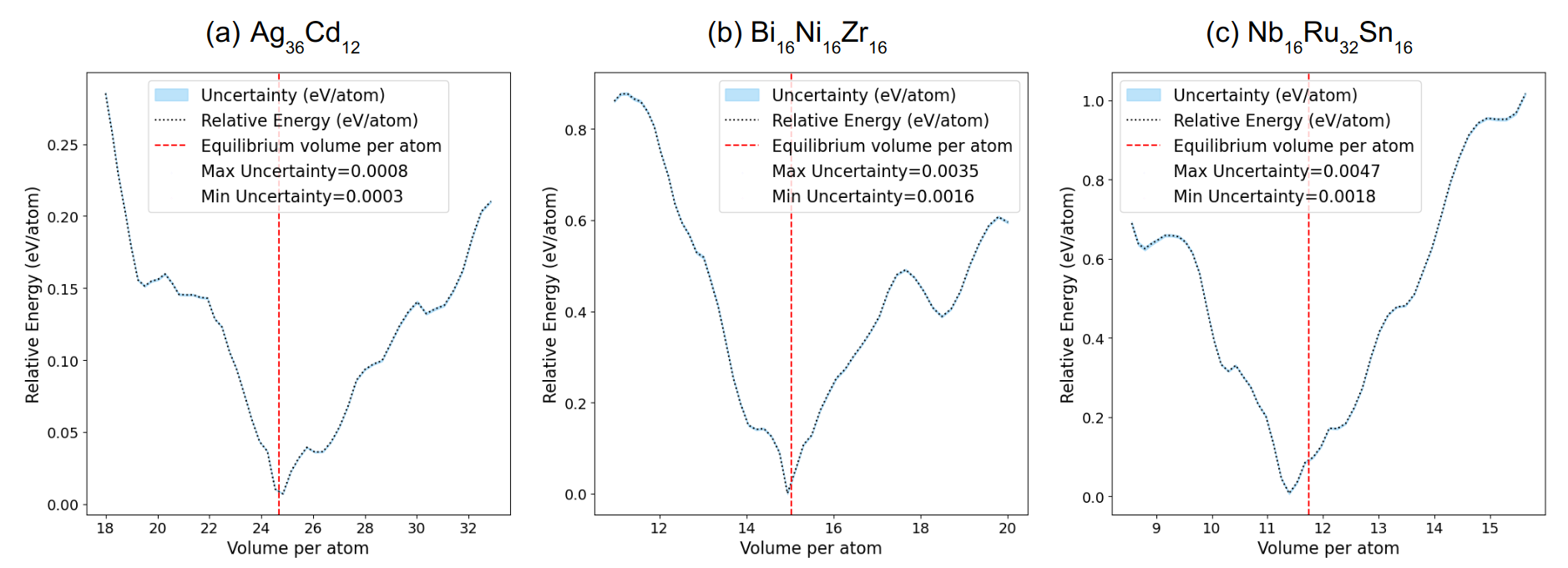}
    \caption{Change in volume per atom for three test cases in inter-metallic slabs: (a) Ag$_36$Cd$_12$, (b) Bi$_16$Ni$_16$Zr$_16$, and (c) Nb$_16$Ru$_32$Sn$_16$. The uncertainty region (shaded in blue) is negligible and overlaps with the relative energy curve (black dotted line), with no discernible trend in uncertainty. The difference between maximum and minimum uncertainty is approximately 0.002 eV/atom, indicating consistently low uncertainty across all cases. The red dotted line is the equilibrium volume per atom for the respective systems. The equilibrium volume per atom, indicated by the red dotted line, aligns with the point of zero relative energy, demonstrating strong model accuracy.}
    \label{fig:metal_slabs}
\end{figure}

To further investigate the model performance, test cases were analyzed for both inter-metallic and nonmetal slabs. The goal was to examine how relative energy and uncertainty vary with changes in volume per atom. For inter-metallic slabs, which are in-domain systems, the model performs as expected. At the equilibrium volume per atom, which is included in the training data, the relative energy is close to zero. Relative energy is defined as the difference between the predicted and actual total energy. As the volume deviates from equilibrium, the relative energy increases, indicating a decline in accuracy. However, there is no trend observed for uncertainty, as the difference between maximum and minimum uncertainty is approximately 0.002 eV/atom, indicating consistently low uncertainty throughout. Fine-tuning the model for inter-metallic slabs made it confident even at non-equilibrium states (see Figure \ref{fig:metal_slabs}).

\begin{figure}
    \centering
    \includegraphics[width=\textwidth]{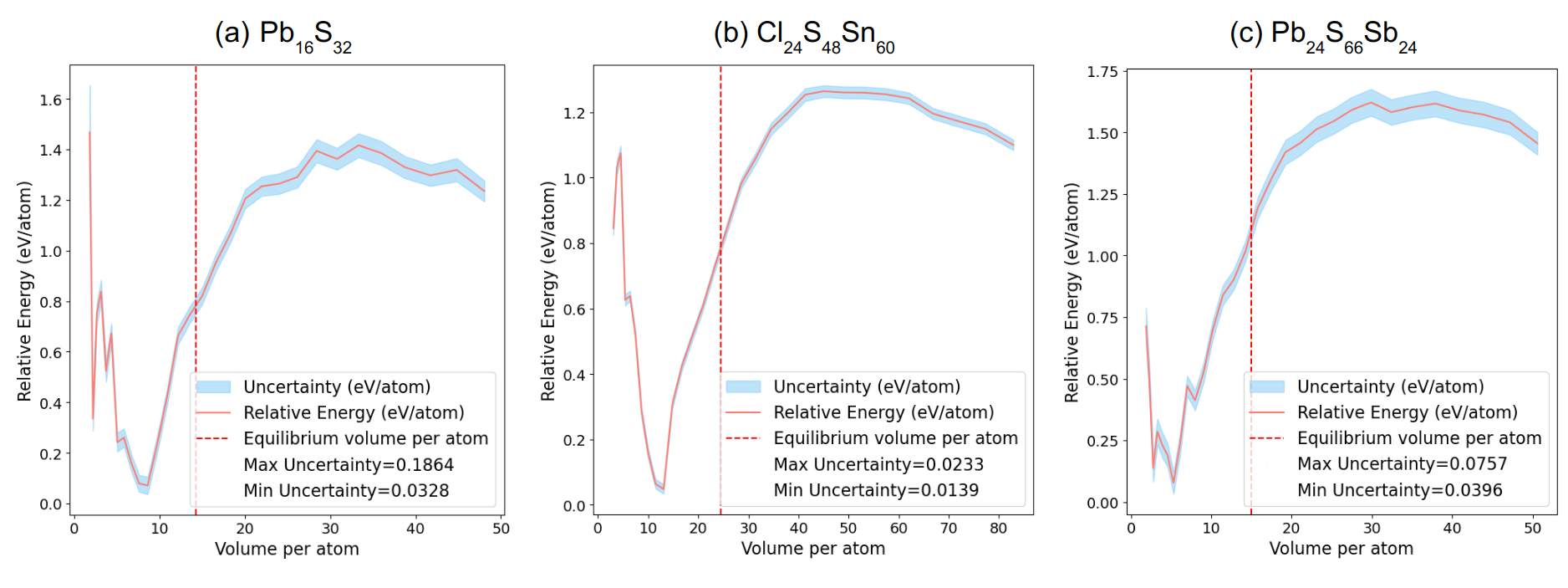}
    \caption{Change in volume per atom for three nonmetal slab systems: (a) Pb$_16$S$_32$, (b) Cl$_24$S$_48$Sn$_60$, and (c) Pb$_24$S$_66$Sb$_24$. The equilibrium volume per atom, indicated by the red dotted line, is not accurately predicted by the model. The relative energy curve (red line) is non-zero at equilibrium, indicating suboptimal model performance. The uncertainty (shaded in blue) follows the expected trend, being lower near equilibrium states and higher away from them.}
    \label{fig:nonmetals}
\end{figure}

Nonmetal slabs, as out-of-domain systems, perform significantly worse than intermetallic slabs due to higher uncertainty values. The relative energy trend deviates from expectations due the model's poor performance. It shows nonzero relative energy at the equilibrium volume per atom (red dotted line in Figure \ref{fig:nonmetals}). For all three test cases (in Figure \ref{fig:nonmetals}), the relative energy decreases below this point before rising again. However, the uncertainty follows the expected pattern. It is lowest near the minimum relative energy and higher away from it.

\subsection{Gold Molecular Dynamics (MD) Dataset}
\subsubsection{Bulk and Amorphous systems}
\begin{figure}
    \centering
    \includegraphics[width=0.9\textwidth]{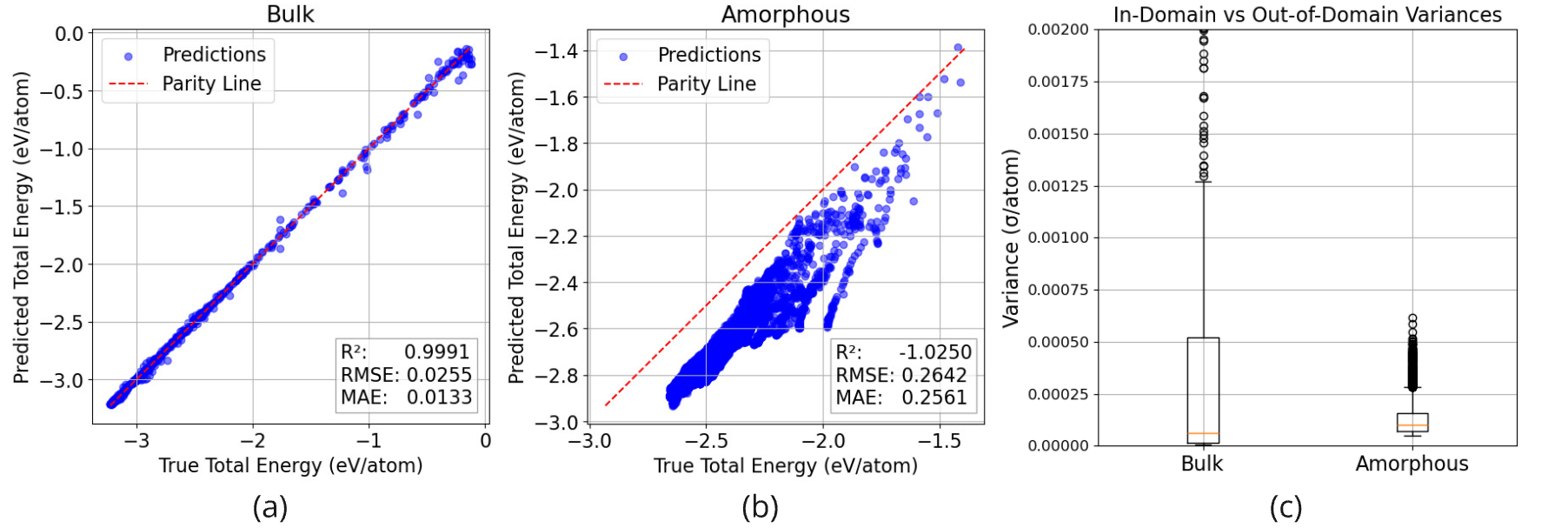}
    \caption{(a) Parity plot for bulk systems showing high accuracy.
    (b) Parity plot for amorphous systems showing lower accuracy.
    (c) Variance estimates showing slightly higher uncertainty for bulk compared to amorphous.}
    \label{fig:gold_model_performance}
\end{figure}

The AuMD-SE model was tested on both bulk and amorphous systems. Results showed higher \(R^2\) scores and lower error values for bulk systems (Figure \ref{fig:gold_model_performance}(a)) compared to amorphous systems, which exhibited lower \(R^2\) and higher errors (Figure \ref{fig:gold_model_performance}(b)). Interestingly, the mean variance was slightly higher for bulk systems, contrary to the expectation of lower uncertainty (Figure \ref{fig:gold_model_performance}(c)). The maximum variance differed by only 0.001 eV/atom, indicating that the uncertainty levels for both systems are similar, given the scale of $10^{-3}$ eV/atom. The model seems to be confident for out-of-domain systems too. 

\begin{figure}
    \centering
    \includegraphics[width=0.9\textwidth]{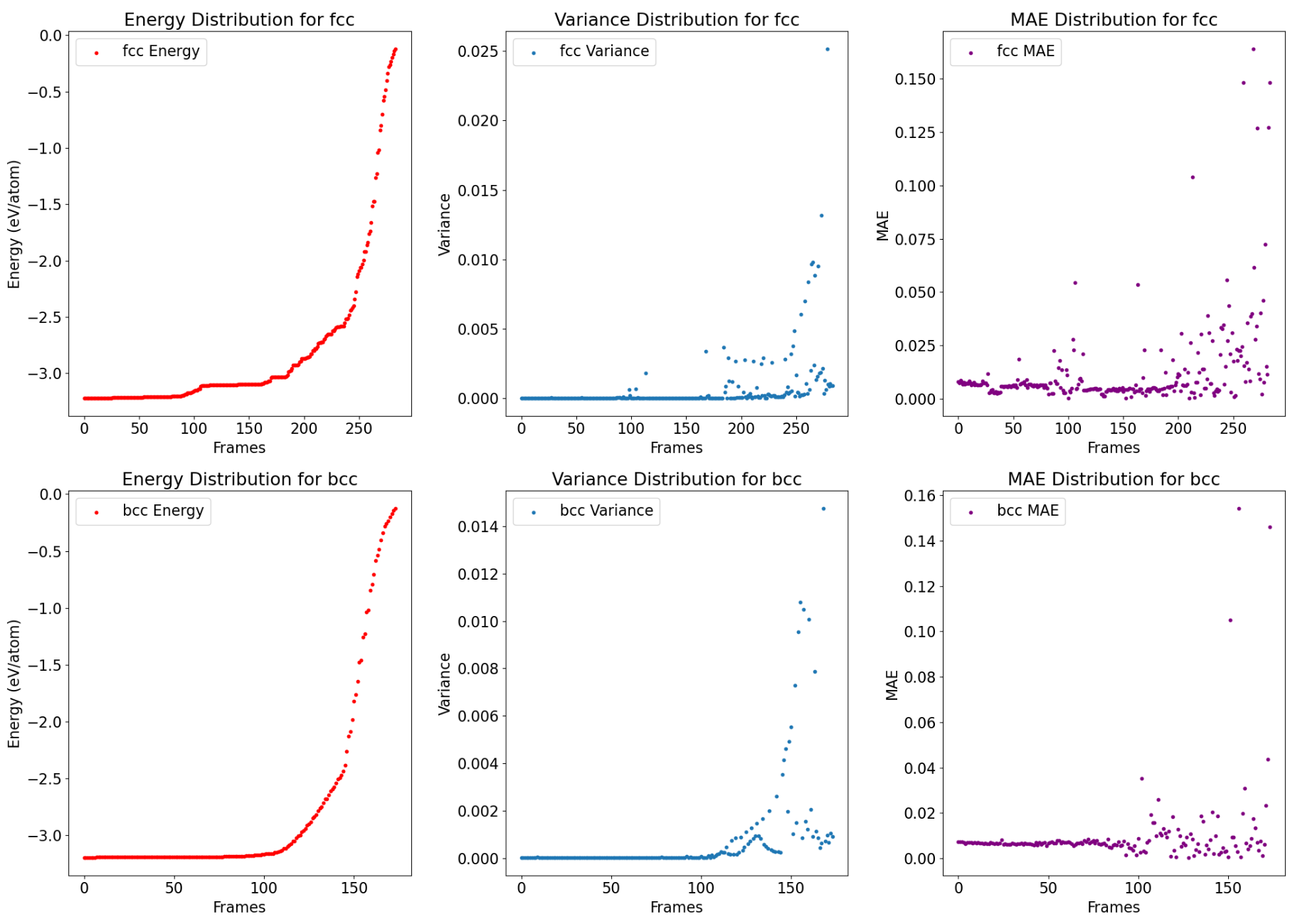}
    \caption{Energy-Variance-Error analysis: Energy, variance, and mean absolute error (MAE) distributions for FCC (top row) and BCC (bottom row) structures. The energy values are arranged in increasing order, with the corresponding variance and error plotted. The results indicate that the model maintains higher confidence at lower energy levels, as reflected by lower variance and error.}
    \label{fig:lattice}
\end{figure}

In this study, energy values are arranged in ascending order and plotted against their corresponding variance and error. A consistent pattern is observed across highly ordered lattices, exemplified by the FCC and BCC structures (as shown in Figure \ref{fig:lattice}). The objective of this analysis is to examine how variance and error evolve with respect to energy. The findings reveal that the model demonstrates higher confidence (lower variance) at lower energy levels, where both variance and error remain minimal. However, as energy increases, the variance also increases, resulting in greater uncertainty in the model’s predictions. This trend indicates that the model's reliability decreases at higher energy states, highlighting the challenges associated with accurately modeling high-energy configurations. This discrepancy arises because the dataset contains a higher number of frames for low-energy states, enabling more accurate modeling, whereas the scarcity of frames for high-energy states results in greater uncertainty and error in predictions.

\begin{figure}
    \centering
    \includegraphics[width=0.9\textwidth]{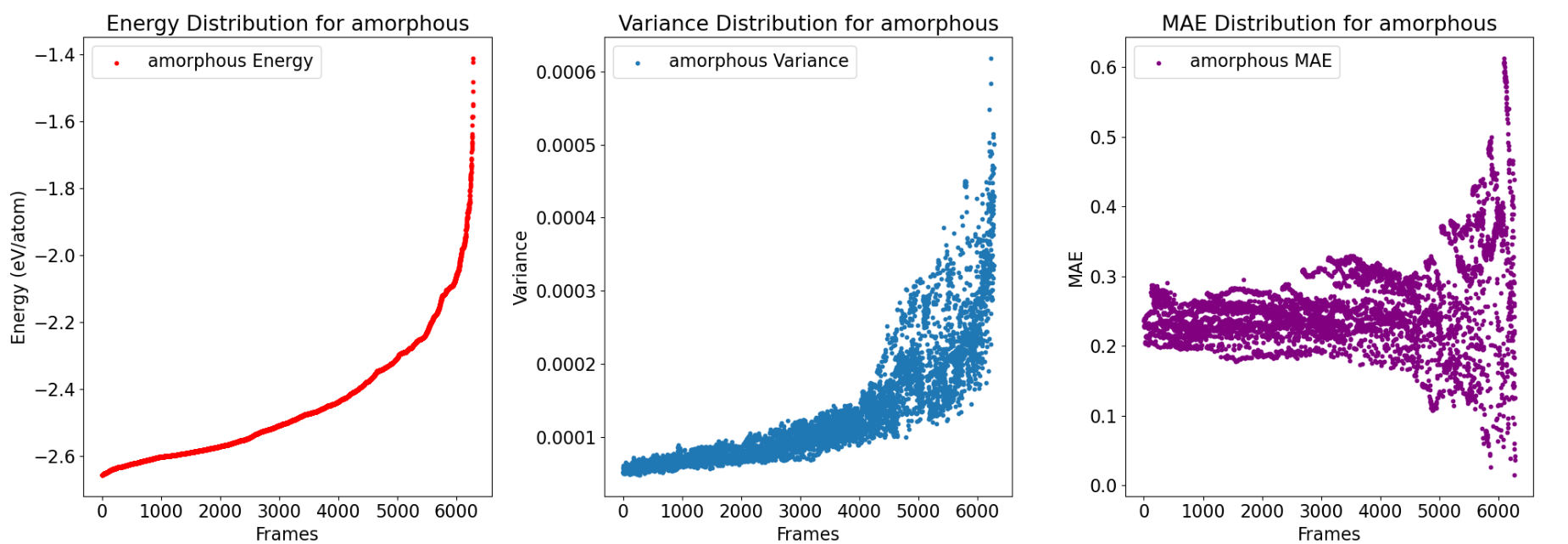}
    \caption{Energy-Variance-Error analysis: Energy, variance, and mean absolute error (MAE) distributions for amorphous structures. The energy values are arranged in increasing order, with the corresponding variance and error plotted. The results indicate that the model maintains higher confidence at lower energy levels, as reflected by lower variance and error.}
    \label{fig:amorphous}
\end{figure}

Similar to highly ordered lattices, amorphous systems also exhibit low uncertainty and error at lower energy values, followed by increasing error and uncertainty at higher energy levels, despite being an out-of-domain dataset (Figure \ref{fig:amorphous}). Notably, the MAE distribution shows a broader spread at higher energy levels, suggesting that the model struggles to maintain accuracy as the system transitions to higher energy states. In fact, the model seems to be more confident about the amorphous systems, as evidenced by the lower variance compared to highly ordered lattices. This can be attributed to two primary factors. First, the presence of short-range order in amorphous systems plays a pivotal role. Even within disordered structures, local atomic arrangements often resemble those found in ordered lattices, such as FCC or BCC configurations. This localized order enables the model to extrapolate patterns and interactions learned from ordered systems to amorphous systems, particularly in low-energy regimes where such order is more evident. Second, the relative uncertainty in predictions for amorphous systems is considerably lower, with variance and error values on the order of \(10^{-4}\) to \(10^{-3} \, \mathrm{eV/atom}\). This reduced uncertainty can be attributed to the fact that amorphous systems near equilibrium exhibit more stable and predictable atomic configurations, whereas ordered lattices span a broader range of energy states, including higher-energy configurations that introduce greater variability. 

\subsubsection{Gold and Silver systems}
Building upon our analysis of the AuMD-SE model’s performance, we further examined its sensitivity to elemental changes by introducing silver (Ag) systems with comparable structures, including bulk, amorphous, and clusters. This approach mirrors our previous evaluations involving inorganic elements absent from the QM9 dataset, where increased uncertainties signified the model’s response to unfamiliar chemical environments.

\begin{figure}
    \centering
    \includegraphics[width=0.5\textwidth]{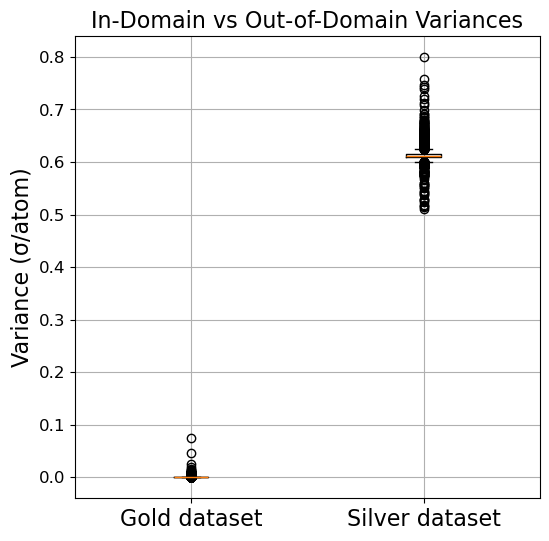}
    \caption{Box plot comparing the variance ($\sigma$/atom) of the AuMD-SE model on in-domain (gold) and out-of-domain (silver) datasets. The gold dataset exhibits low variance, indicating high confidence due to fine-tuning on gold systems. In contrast, the silver dataset shows significantly higher variance, highlighting the model's ability to capture out-of-domain characteristics.}
    \label{fig:au_ag}
\end{figure}

Initially, the AuMD-SE model was exclusively trained on gold (Au) systems, yielding consistently low uncertainties due to extensive exposure during fine-tuning. However, introducing silver systems resulted in significantly higher uncertainty levels. Specifically, the mean variance observed for gold was exceptionally low at 0.00022 eV/atom, whereas for silver, the variance substantially increased to 0.61304 eV/atom (Figure \ref{fig:au_ag}).

This notable variance difference highlights the model’s distinct sensitivity to elemental composition changes. In contrast to the minor uncertainty difference observed between bulk and amorphous gold systems, the marked increase for silver clearly demonstrates the model’s effectiveness in identifying and quantifying uncertainties for elements outside its trained domain. These findings reinforce the robustness of the AuMD-SE model in recognizing novel chemical contexts, aligning with our earlier conclusions drawn from the QM9 dataset analysis.

\section{Conclusion}
This study explored Uncertainty Quantification (UQ) in Graph Neural Networks (GNNs) using the Direct Propagation of Shallow Ensembles (DPOSE) approach. By integrating DPOSE into SchNet, a widely used model for atomistic simulations, we demonstrated its effectiveness in providing reliable uncertainty estimates while maintaining computational efficiency. Through evaluations on the QM9, OC20, and Gold datasets, our results indicate that DPOSE successfully distinguishes in-domain from out-of-domain data when they are far apart, by exhibiting big differences in uncertainty. Furthermore, the model accurately captured uncertainty trends related to molecular size, structural distortions, and elemental composition variations.

For QM9, uncertainty increased for molecules with elements absent in training and larger molecular systems, confirming the method's sensitivity to domain shifts. In the OC20 dataset, the model showed confidence in inter-metallic slab predictions while highlighting higher uncertainties for non-metal systems, aligning with expected model limitations. Similarly, in the Gold dataset, the energy-variance-error analysis revealed consistent trends in ordered and amorphous structures, emphasizing the model’s ability to recognize high-energy states as more uncertain. However, the DPOSE framework struggled to distinguish between in-domain and out-of-domain configurations in this dataset, perhaps indicating they are not as different as we expected. Consequently, differentiating these equilibrium amorphous systems from in-domain bulk systems proved difficult. Nevertheless, similar to the observations in QM9 dataset with out-of-domain compositions, DPOSE showed a clear distinction between in-domain and out-of-domain data for the Gold and Silver datasets. This indicates that the model effectively captures differences when atom types change but struggles to differentiate structurally similar systems with the same atom type. Despite these challenges, DPOSE remains a computationally efficient and scalable solution for uncertainty quantification in GNNs, contributing to enhanced reliability in materials discovery.

\section*{References}
\bibliographystyle{iopart-num}
\bibliography{references}

\end{document}